\pdfoutput=1
\documentclass[runningheads]{llncs}
\usepackage{makecell}
\usepackage[utf8]{inputenc}
\usepackage{csquotes}
\usepackage{amsmath}
\usepackage{amssymb,amsfonts}
\usepackage{pifont}
\usepackage{algorithmic,algorithm}
\usepackage{mathtools}\usepackage{todonotes}
\usepackage{url}
\usepackage{subfig}
\usepackage[para,online,flushleft]{threeparttable}
\usepackage{booktabs}
\usepackage{graphicx}
\usepackage{xcolor}
\usepackage{multirow}
\usepackage{booktabs}
\usepackage{bbding}
\usepackage[
    style=ieee,
    citestyle=numeric-comp,
    backend=biber,
    hyperref=true,
    maxnames=1,
    maxbibnames =3,
    minnames=1,
    mincrossrefs=1,
    natbib=true,
    sorting=none,
    sortcites=true,
    url=false,
    doi=false,
    eprint=false
]{biblatex}
\AtEveryBibitem{%
  \clearname{translator}%
  \clearlist{publisher}%
  \clearfield{pagetotal}%
\clearfield{pagetotal}%
\clearfield{issn}%
\clearfield{editor}%
\clearfield{month}%
\clearfield{notes}%
\clearfield{language}%

}
\DeclareFieldFormat
  [article,inbook,incollection,inproceedings,patent,thesis,unpublished]
  {title}{{#1}}

\usepackage[english]{babel}
\usepackage[colorlinks=true,urlcolor= blue]{hyperref}
\def\@fnsymbol#1{\ensuremath{\ifcase#1\or *\or \dagger\or \ddagger\or
   \mathsection\or \mathparagraph\or \|\or **\or \dagger\dagger
   \or \ddagger\ddagger \else\@ctrerr\fi}}
\makeatletter
\newcommand{\ssymbol}[1]{^{\@fnsymbol{#1}}}
\makeatother

\newcommand{\Loss}{\mathcal{L}}
\newcommand{\Reals}{\mathbb{R}}

\newcommand{\atlas}{\mathbf{y}_a}
\newcommand{\prediction}{\hat{\mathbf{y}}}
\newcommand{\transform}{\mathbf{\Phi}}
\newcommand{\feature}{\mathbf{f}}

\newcommand{\x}{\mathbf{x}}
\newcommand{\y}{\mathbf{y}}

\newcommand{\attention}{\mathbf{a}}
\newcommand{\W}{\mathbf{W}}
\newcommand{\bias}{\mathbf{b}}
\newcommand{\red}[1]{\textcolor{red}{#1}}%
\newcommand{\segNet}{\mathcal{S}}

\newcommand{\atlasNet}{\mathcal{A}}
\newcommand{\datahref}[2]{\href{#1}{\textcolor{black}{#2}}}%
\newcommand\blfootnote[1]{%
  \begingroup
  \renewcommand\thefootnote{}\footnote{#1}%
  \addtocounter{footnote}{-1}%
  \endgroup
}

\begin{document}
\title{Pay Attention to the Atlas: Atlas-Guided Test-Time Adaptation Method for Robust 3D Medical Image Segmentation}

\titlerunning{AdaAtlas for Robust 3D Medical Image Segmentation}
\author{Jingjie Guo\inst{1}(\Envelope), Weitong Zhang~\inst{4}, Matthew Sinclair\inst{3,4}, Daniel Rueckert\inst{1,2,4}, Chen Chen\inst{4,5,6}}
\authorrunning{J. Guo et al.}
\institute{School of Computation, Information and Technology, Technical University of Munich, Germany
\and School of Medicine, Klinikum rechts der Isar, Technical University of Munich, Germany
\and HeartFlow, Inc., California, USA
\and Department of Computing, Imperial College London, UK
\and Department of Engineering Science, University of Oxford, UK 
\and Department of Computer Science, University of Sheffield, UK \\
\email{jingjie.guo@tum.de
}}

\maketitle            
\begin{abstract}
\blfootnote{This work was partly done at HeartFlow, Inc.}
Convolutional neural networks (CNNs) often suffer from poor performance when tested on target data that differs from the training (source) data distribution, particularly in medical imaging applications where variations in imaging protocols across different clinical sites and scanners lead to different imaging appearances. However, re-accessing source training data for unsupervised domain adaptation or labeling additional test data for model fine-tuning can be difficult due to privacy issues and high labeling costs, respectively. To solve this problem, we propose a novel atlas-guided test-time adaptation (TTA) method for robust 3D medical image segmentation, called AdaAtlas. AdaAtlas only takes one single unlabeled test sample as input and adapts the segmentation network by minimizing an atlas-based loss. Specifically, the network is adapted so that its prediction after registration is aligned with the learned atlas in the atlas space, which helps to reduce anatomical segmentation errors at test time. In addition, different from most existing TTA methods which restrict the adaptation to batch normalization blocks in the segmentation network only, we further exploit the use of channel and spatial attention blocks for improved adaptability at test time. Extensive experiments on multiple datasets from different sites show that AdaAtlas with attention blocks adapted (AdaAtlas-Attention) achieves superior performance improvements, greatly outperforming other competitive TTA methods. 
\end{abstract}

\section{Introduction}
Convolutional neural networks have shown good results in medical image segmentation~\cite{ronneberger2015u}. However, domain shifts between the training and test data often lead to significant performance degradation, which is a common problem for medical image segmentation since medical images can be acquired by devices from different companies using different scanning protocols in different medical centers. These differences can lead to distribution shifts of the data, causing the model to perform poorly on images from unseen target domains. Directly labeling target domain data for model fine-tuning is not feasible due to high labeling costs. To mitigate this issue, researchers have proposed various approaches such as unsupervised domain adaptation~\cite{Tao_2019_Radiology,liu2020ms,liu2020saml,Dou_2019_NIPS} and domain generalization~\cite{chen2022maxstyle, Wang_IJCAI_2021_DG_Survey,Xu_2021_ICLR_RandConv,Chen_2020_MICCAI_Adv_Bias, Chen_2021_latent_space_data_augmentation}. However, most unsupervised domain adaptation methods require access to data from target domains at training time, which can be hard to get due to privacy issues. On the other hand, most domain generalization methods require multi-domain source data, which can be challenging since sharing medical images is undesired. Also, the learned `generalized' solution can still be sub-optimal for a specific target domain with unique characteristics. Different from them, test-time adaptation (TTA) methods aim to solve this problem by adapting and optimizing the trained model to each specific domain, with access to target domains only at test time.
Since there is no label information available at test time, different surrogate losses have been proposed to guide the model adaptation~\cite{sun2020test, fleuret2021test, wang2020tent, niu2022efficient,bateson2022test, bateson2022source, Huang_MICCAI2022_online,karani2021test,liu2022single}. Many of them focus on utilizing pixel-level information~\cite{sun2020test, fleuret2021test, wang2020tent, niu2022efficient}, e.g. entropy loss~\cite{wang2020tent,niu2022efficient}. Yet, for segmentation tasks, such information can be limited as it does not consider global shape information. Also, we argue that entropy loss-based TTA such as TENT can be unstable as deep neural networks are known to be over-confident. Another common limitation of existing approaches is that they limit the adaptation to normalization blocks i.e., batch normalization in order to stabilize the adaptation process. The adaptability can, however, be restricted, as they only allow for the rescaling and shifting of the features in the channel dimension.


To solve the above limitations, we propose a novel TTA method called AdaAtlas, which uses an atlas, a high-level shape prior for more reliable adaptation at test time. We further consider incorporating attention blocks for higher model adaptability at test time (AdaAtlas-Attention), which allows adapting features in both channel-wise and spatial-wise manner. Of note, while prior works have developed different attention blocks to improve model representation learning capacity, their effectiveness has mostly been evaluated on the source domain whereas their capacity for test-time adaptation on unseen target domains is yet under-explored. Our contribution can be listed as follows: a) We propose a novel atlas-based method for medical image segmentation, which considers 3D shape prior information for reliable test-time adaptation  (Sec.~\ref{sec:atlas_TTA}); b) To the best of our knowledge, we are the first to consider employing attention blocks for improved adaptability at test time (Sec.~\ref{sec:ada_blocks}). Our results show that adapting attention blocks rather than normalization blocks can lead to better performance especially when big distribution shifts are exhibited; c) We perform extensive experiments on multiple datasets from different sites, and show that our method can substantially increase the performance on target domains and outperforms other competitive TTA methods by a large margin (Sec.~\ref{sec:results}). 

\noindent\textbf{Related work.}
\textbf{\textit{a) TTA for image segmentation}}: TTA adapts a source-domain pre-trained model at test time to a specific target domain via minimizing certain loss objectives for improved performance without ground truth provided. Many existing works employ unsupervised loss, such as entropy-based loss~\cite{wang2020tent, niu2022efficient}, or self-supervised loss with pre-defined auxiliary tasks~\cite{sun2020test}. Yet, most of them operate at the pixel level, which can be limited to guide high-level image segmentation models. To address the above limitation, methods that distill higher-order information from the source domain for better supervision have been proposed, including class-ratio prior~\cite{bateson2022source}, shape dictionary~\cite{liu2022single}, shape moment-based loss~\cite{bateson2022test}, or pseudo anatomical labels generated by denoising autoencoder (DAE)~\cite{karani2021test}. In particular, the DAE-based approach~\cite{karani2021test} needs to train a separate DAE to produce `corrected' segmentation to provide supervision and adapt another image normalization network to adjust the input test image so that the fixed segmentation network can produce anatomically plausible segmentation on the adapted test image. Similarly, our work utilizes a source-domain learned atlas as high-level shape prior but computes the loss in the atlas space. Compared to DAE, we adapt the segmentation network, which does not alter the information in the test image. \textbf{\textit{b) Atlas-based segmentation}}: Recent efforts have been made to boost the segmentation performance by combining existing deep segmentation networks with an atlas registration network via joint training, and performing atlas-to-subject registration to get a warped atlas as the final prediction for a subject~\cite{dong2020deep, sinclair2022atlas}. Yet, we found that their methods can easily fail when the target segmentation for a subject has lots of false positives or totally missed predictions. Thus, in this work, we focus on directly improving the quality of predicted segmentation via test-time adaptation, where the atlas is fixed and treated as a reliable shape prior to guide the adaptation of the segmentation network at test time.

\section{Method}
\subsection{Atlas-guided test-time adaptation framework (AdaAtlas)}
\label{sec:atlas_TTA}
\begin{figure}[t]
    \centering   \includegraphics[width=\textwidth]{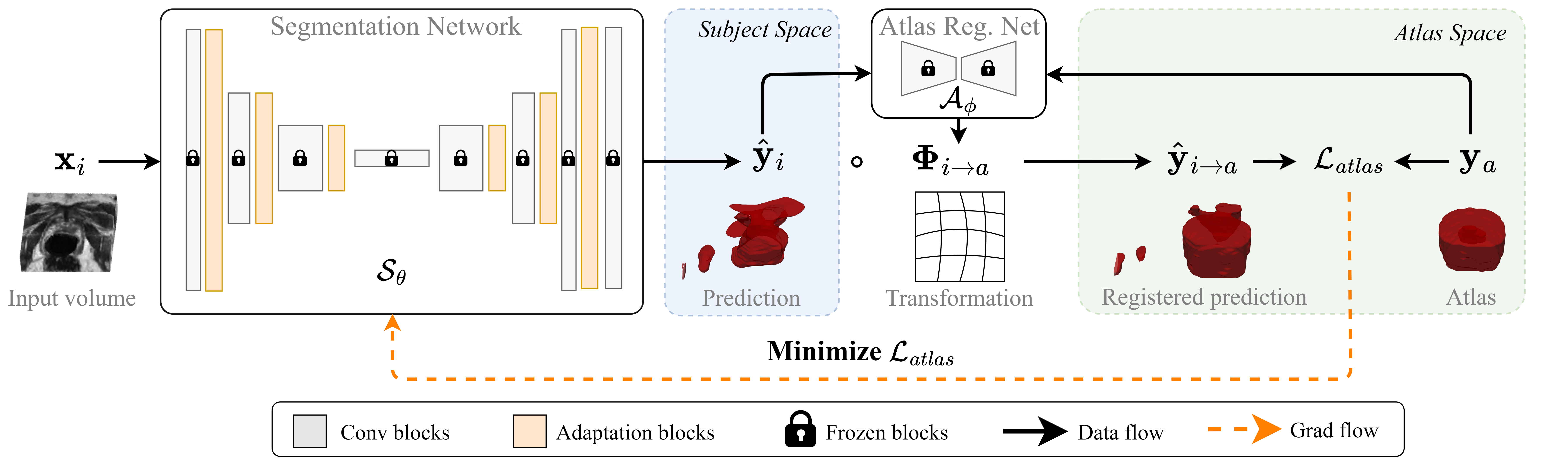}
    \caption{\textbf{AdaAtlas overview:} Our TTA method utilizes an atlas-based shape prior loss $\Loss_{atlas}$ (Eq.\ref{eq:atlas_loss}) to adapt $\segNet_\theta$ on each \textbf{\textit{single}} subject $\boldsymbol{x}_i$, which encourages the prediction $\prediction_i$ to be aligned with a given atlas $\atlas$ after registration (reg.) in the atlas space (Sec.~\ref{sec:atlas_TTA}). For efficiency, during TTA, only parts of the segmentation network, i.e., adaptation blocks (orange blocks, details in Sec.~\ref{sec:ada_blocks}) are updated.}
    \label{fig:AdaAtlas_overview}
\end{figure}
As shown in Fig.~\ref{fig:AdaAtlas_overview}, our proposed method first uses a 3D segmentation network $\segNet_\theta$ to take an image $\x_i \in \Reals^{1\times H \times W \times D}$ as input and predicts a class-wise probabilistic segmentation map $\prediction_i \in \Reals^{C \times H \times W \times D}$ with $C$ being the number of predicted classes. A 3D atlas registration (atlas reg.) network $\atlasNet_\phi$ then takes the predicted segmentation $\prediction_i$ and a learned 3D atlas $\atlas  \in \Reals^{C \times H \times W \times D}$ as input and predicts a deformation field $\transform \in \Reals^{3\times H \times W \times D}$ (affine+non-rigid transformation) that registers the predicted map to the atlas. Of note, the segmentation network and the atlas registration network are pre-trained with data from the source domain~\cite{sinclair2022atlas}. The 3D atlas $\atlas$ is constructed and refined iteratively during training. Specifically, the atlas is first initialized as an average of training samples' labels. It is then updated by warping training labels to the atlas space using the registration network and averaging again across all samples using an exponential moving average at the end of each training epoch. The two networks are jointly optimized by performing segmentation, atlas-to-segmentation registration, and segmentation-to-atlas registration tasks simultaneously. By continuously feeding network predictions to the registration network across all training subjects to learn mappings between subjects and the probabilistic atlas, the registration network is enhanced to capture a wide range of anatomical variations and their correspondence to the atlas in the unified atlas space. As the main focus of this paper is about test-time adaptation, we refer interested readers to \cite{sinclair2022atlas} for more training details.

At test time, given a test image $\x_i$ from an unseen target domain, our goal is to adapt the segmentation network $\segNet_\theta$ so that the predicted segmentation is improved. Since there is no ground truth to finetune the model on the target domain, we utilize the atlas $\atlas$ as a high-level shape prior to guiding the adaptation. As the atlas is used to represent the standardized view of the anatomical structure, we believe that it can be used as a domain-invariant concept to enable reliable adaptation at test time. Specifically, given the prediction $\prediction_i$ from the pre-trained segmentation network for test image $\x_i$  and the learned atlas $\atlas$, we feed both into the pre-trained atlas registration network $\atlasNet_\phi$ (frozen at test time) to estimate the deformation $\transform = \atlasNet_\phi(\prediction_i,\atlas)$ that transforms the prediction from the subject space to the atlas space 
$ {\prediction}_{i\rightarrow a} = \prediction_i \circ \transform
$. We then measure the difference between the registered prediction ${\prediction}_{i\rightarrow a}$ and the reliable atlas $\atlas$ in the same space and use that as a loss to adapt the segmentation network. The loss $\Loss_{atlas}$ to optimize $\segNet_\theta$ is defined as: 
\begin{equation}
\label{eq:atlas_loss}
\Loss_{atlas}({\prediction}_{i\rightarrow a}, \atlas) = 1 - \frac{1}{H\times W \times D} \sum^H_{h=1}\sum^W_{w=1}\sum^D_{d=1} \cos({\prediction}_{i\rightarrow a}^{h,w,d}, \atlas^{h,w,d}).
\end{equation}
Here $\cos(\cdot, \cdot)$ is the commonly used cosine function for similarity measurement.

\subsection{Where to adapt: adaptation blocks in segmentation network}
\label{sec:ada_blocks}

\begin{figure}[t]
    \centering   
    \includegraphics[width=0.9\textwidth]{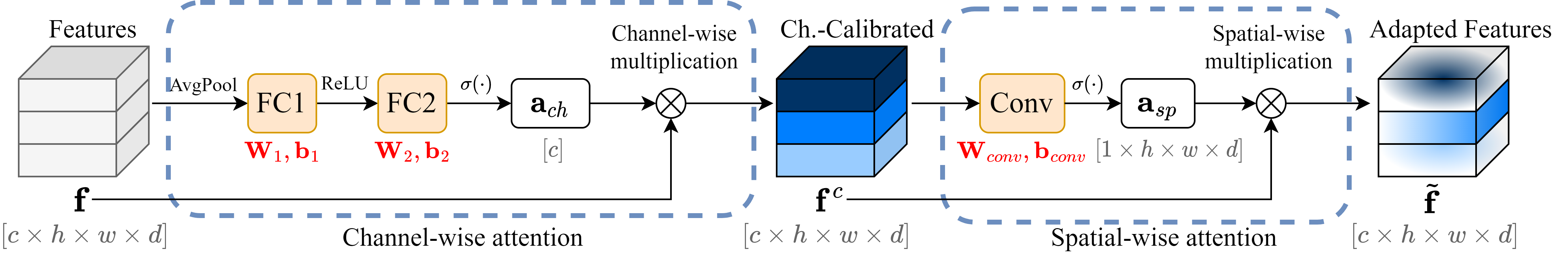}
    \caption{Dual attention blocks with test-time adaptable parameters in \textcolor{red}{red}.}
    \label{fig:adaptation_blocks}
\end{figure}

Instead of optimizing all the parameters in $\segNet_\theta$ to minimize $\Loss_{atlas}$, it is more common to only adapt part of the segmentation network for improved efficiency and effectiveness~\cite{wang2020tent}. A commonly adopted method is adapting scaling and shifting parameters in the batch normalization blocks~\cite{wang2020tent,niu2022efficient,bateson2022source,bateson2022test}. Yet, this type of adaptation only allows channel-wise feature calibration. Inspired by the success of attention-based works~\cite{oktay2018attention,roy2018concurrent, woo2018cbam,guo2022attention}, we design a dual attention block that can calibrate the features in both channel-wise and spatial-wise fashion, with the goal to improve the adaptation power. These attention blocks can be inserted into any CNN architecture. As shown in Fig.~\ref{fig:adaptation_blocks}, given a feature map $\feature$, a dual attention block first computes a channel-wise attention score $\attention_{ch} \in \Reals^c$, which recalibrates $\feature$ to $\feature^c$ via channel-wise multiplication. The intermediate output $\feature^c$ is then sent to the spatial attention module to compute spatial-wise attention score $\attention_{sp} \in \Reals^{1 \times h \times w \times d}$ to recalibrate features via spatial-wise multiplication. In practice, we found that sequential dual attention works better than the existing concurrent dual attention in \cite{roy2018concurrent}. At a high level, it is defined as: $\tilde{\feature} = \attention_{sp} \star ( \attention_{ch} \times \feature).$ Here $\times$ represents channel-wise multiplication and $\star$ represents spatial-wise multiplication. To compute $\attention_{ch}$ for calibrating $\feature$, we first flatten the feature $\feature$ via average pooling and then employ two fully connected layers (FC1, FC2) and a sigmoid function to get a channel-wise score in $[0,1]$. To compute $\attention_{sp}$, we employ a 3D $1\times 1 \times 1$ convolutional layer (Conv) followed by a sigmoid function to process intermediate channel-wise calibrated feature $\feature^c$ to final adapted feature $\tilde{\feature}$. Different from prior work which fixes the kernels of attention blocks at test time, we revolutionize its use for TTA. At test time, the parameters in the dual attention blocks (incl. weights $\red{\W_{1}} \in \Reals^{\frac{c}{r} \times c}$, $\red{\W_{2}} \in \Reals^{c \times \frac{c}{r}}$, $\red{\W_{conv}} \in \Reals^{1\times 1\times 1\times c}$ and biases $\red{\bias_1} \in \Reals^{\frac{c}{r}}$,$\red{\bias_2} \in \Reals^{c}$, $\red{\bias_{conv}} \in \Reals$ in two FC layers and the Conv layer, respectively) are updated in the direction of minimizing the $\Loss_{atlas}$. 

\section{Experiments and results}

\begin{figure}[t]
    \centering   
    \includegraphics[width=0.8\textwidth]{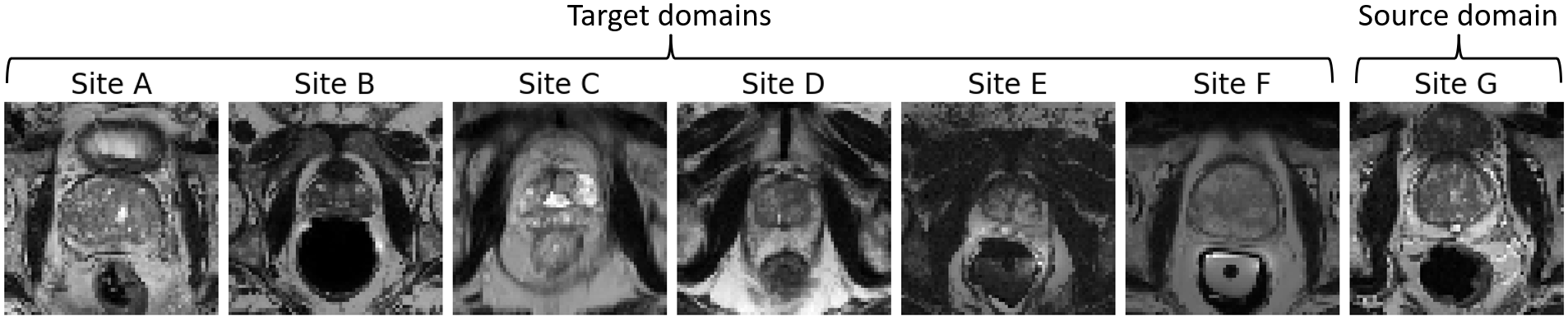}
    \caption{Visualization of prostate MRI datasets from different sites. Sites A-F are target domains for testing. Site G is the training source domain.}
    \label{fig:datasets_visual}
\end{figure}

 
\textbf{Data: prostate multi-site segmentation datasets.} To verify the effectiveness of our algorithm, we evaluate our method on the prostate segmentation task from T2-weighted MRI. Specifically, we use multi-site prostate segmentation datasets (collated from four public datasets \datahref{http://medicaldecathlon.com/}{MSD}~\cite{antonelli2022medical}, \datahref{https://wiki.cancerimagingarchive.net/display/Public/NCI-ISBI+2013+Challenge+-+Automated+Segmentation+of+Prostate+Structures}{NCI-ISBI13}~\cite{bloch2015nci}, \datahref{https://i2cvb.github.io/}{I2CVB}~\cite{lemaitre2015computer} and \datahref{https://promise12.grand-challenge.org/}{PROMISE12}~\cite{litjens2014evaluation}). They consist of 148 images from \textbf{seven, different} clinical sites. All subjects are uniformly resized to $64 \times 64 \times 64$ and normalized to have zero mean and unit variance in intensity values. We use the single-source MSD dataset (G) as the source domain (22 subjects for training), while the remaining images from the other \textbf{six, unseen} clinical sites (A-F: 30/30/19/13/12/12 images) are used for testing, see Fig~\ref{fig:datasets_visual}.

\noindent\textbf{Implementation and baselines.}  We conducted experiments using U-Net~\cite{ronneberger2015u} and its variant with additional inserted dual attention blocks after each convolution block in the segmentation network except for the last layer, as suggested by~\cite{roy2018concurrent}. We used the framework described in \cite{sinclair2022atlas} to train the segmentation and the atlas registration network jointly and construct the atlas on the source domain. The training loss consists of a standard supervised segmentation loss and a bidirectional registration loss $\Loss_{bireg}$, as well as a regularization term to encourage the smoothness of the predicted deformation field. The registration loss computes the dissimilarity between the atlas and the predicted segmentation in both the subject space and the atlas space \cite{sinclair2022atlas}: $\Loss_{bireg} = \Loss_{reg}(\prediction_i, {\y}_{a\rightarrow i}) + \Loss_{reg}(\atlas, \prediction_{i\rightarrow a})$. We trained the networks for 800 epochs with a batch size of 8. Adam optimizer was used with an initial learning rate of 0.001 followed by an exponential decay with a half-life of 400 epochs. At test time, we used the pre-trained model as model initialization and employed Adam optimizer (learning rate=0.001) and $\Loss_{atlas}$ to update the adaptation blocks (norm/dual-attention) for 50 iterations for each single test subject, which took about 20 seconds. Of note, we use `\emph{Baseline}' to denote the source domain trained models without any adaptation. The code for AdaAtlas can be found at \href{https://github.com/jingjieguo/AdaAtlas}{github.com/jingjieguo/AdaAtlas}.
    
For fairness and ease of comparison, if not specified, we used the same pre-trained segmentation model (U-Net or U-Net with dual attention blocks) described above for model initialization to compare different TTA methods. Specifically, we compared our method to popular post-hoc TTA methods: TENT~\cite{wang2020tent}, which uses entropy loss to optimize norm blocks; three TENT enhanced variants with modified losses for stronger supervision (EATA using confidence threshold-based entropy loss ~\cite{niu2022efficient}, AdaMI with class-ratio prior~\cite{bateson2022source}, and TTAS with shape-moment-based loss~\cite{bateson2022test}). We further compared our method to three TTA methods that required additional training modifications: a) TTT~\cite{sun2020test}, which requires attaching a self-supervised image reconstruction decoder to the segmentation network at training and test time to construct a self-supervision signal for updating the shared encoder in the segmentation network; b) DAE~\cite{karani2021test}, which requires to train an additional denoising autoencoder to generate `corrected' pseudo labels, and train and adapt an additional image normalization network to adjust the test image instead of the segmentation network at test time; c) TTR~\cite{sinclair2022atlas}, which shares the same source domain training setting as ours. Differently, it refines the registration network at test time (and not the segmentation network) to get an optimized deformed atlas as the final prediction. All methods were implemented in PyTorch with their recommended TTA setups.

\begin{table}[t]
\centering
\caption{Segmentation results of different test-time adaptation methods on \textbf{six, unseen} prostate MRI domains. Reported values are average Dice scores (0: mismatch, 1: perfect match).}
\label{tab:prostate_TTA}
\resizebox{0.9\textwidth}{!}
{%
\begin{threeparttable}
\setlength{\tabcolsep}{3pt}
\begin{tabular}{c|c|llllll|c}
\hline
Backbone                                        & Method                   & \multicolumn{1}{c}{A} & \multicolumn{1}{c}{B} & \multicolumn{1}{c}{C} & \multicolumn{1}{c}{D} & \multicolumn{1}{c}{E} & \multicolumn{1}{c|}{F}      & Mean (A-F)  \\ \hline
\multirow{9}{*}{U-Net}                          & \emph{Baseline}                 & 0.8242                & 0.6025                & 0.5598                & 0.7522                & 0.5620                & 0.6653                      & 0.6610~\textcolor{gray}{(-)} \\ \cline{2-9} 
                                                & TENT                     & 0.8303                & 0.6148                & 0.5630                & 0.7678                & 0.5849                & 0.6753                      & 0.6727~\textcolor{gray}{(+2\%)} \\
                                                & EATA                     & 0.8441                & 0.6207                & 0.5643                & 0.7628                & 0.6349                & 0.7094                      & 0.6894~\textcolor{gray}{(+4\%)} \\
                                                & AdaMI                    & 0.8446                & 0.6913                & 0.6710                & 0.8198                & 0.6613                & 0.7454                      & 0.7389~\textcolor{gray}{(+12\%)} \\
                                                & TTAS                     & 0.8486                & 0.6987                & 0.6956                & \textbf{0.8307}                & 0.6673                & 0.7649                      & 0.7510~\textcolor{gray}{(+14\%)} \\ \cline{2-9} 
                                                & TTT                      & 0.8275                & 0.6175                & 0.5608                & 0.7716                & 0.5786                & 0.6684                      & 0.6707~\textcolor{gray}{(+1\%)} \\
                                                & DAE                      & 0.8494                & 0.6872                & 0.6618                & 0.7828                & 0.5645                & 0.7733                      & 0.7198~\textcolor{gray}{(+9\%)} \\
                                                & TTR                      & 0.8548                & 0.7006                & 0.6889                & 0.7872                & 0.6198                & 0.7430                      & 0.7324~\textcolor{gray}{(+11\%)} \\ \cline{2-9} 
                                                & \textbf{AdaAtlas-Norm(ours)  }    & \textbf{0.8577}                & \textbf{0.7856}                & \textbf{0.7190}                & 0.8258                & \textbf{0.7280}                & \textbf{0.8057}                      & \textbf{0.7870~\textcolor{gray}{(+19\%)}}* \\ \hline \hline
\multirow{10}{*}{\makecell{U-Net + \\ \textit{Dual Attention Blocks}}} & \emph{Baseline}                 & 0.8302                & 0.6072                & 0.5764                & 0.7737                & 0.5751                & 0.6994                      & 0.6770~\textcolor{gray}{(-)} \\ \cline{2-9} 
                                                & TENT                     & 0.8405                & 0.6290                & 0.5876                & 0.7939                & 0.6002                & 0.7083                      & 0.6933~\textcolor{gray}{(+2\%)} \\
                                                & EATA                     & 0.8506                & 0.6636                & 0.6230                & 0.7946                & 0.6203                & 0.7133                      & 0.7109~\textcolor{gray}{(+5\%)} \\
                                                & AdaMI                    & 0.8795                & 0.7101                & 0.6899                & 0.8348                & 0.6783                & 0.7502                      & 0.7571~\textcolor{gray}{(+12\%)} \\
                                                & TTAS                     & 0.8823                & 0.7372                & 0.7270                & 0.8399                & 0.6771                & 0.7658                      & 0.7716~\textcolor{gray}{(+14\%)} \\ \cline{2-9} 
                                                & TTT                      & 0.8385                & 0.6188                & 0.5784                & 0.8045                & 0.5948                & 0.7045                      & 0.6899~\textcolor{gray}{(+2\%)} \\
                                                & DAE                      & 0.8837                & 0.7050                & 0.7084                & 0.8036                & 0.5807                & 0.7685                      & 0.7417~\textcolor{gray}{(+10\%)} \\
                                                & TTR                     & 0.8754                & 0.7446                & 0.6672                & 0.8474                & 0.6123                & 0.7514                      & 0.7497~\textcolor{gray}{(+11\%)} \\ \cline{2-9} 
                                                & \textbf{AdaAtlas-Norm(ours)    }  & 0.8815                & 0.7986                & 0.7694                & 0.8458                & 0.7294                & 0.8075                      & 0.8054~\textcolor{gray}{(+19\%)}* \\
                                                & \textbf{AdaAtlas-Attention(ours)} & \textbf{0.8852}                & \textbf{0.8217}                & \textbf{0.8133}                & \textbf{0.8482}                & \textbf{0.7389}                & \multicolumn{1}{c|}{\textbf{0.8156}} & \textbf{0.8205~\textcolor{gray}{(+21\%)}}* \\ \hline
\end{tabular}

\begin{tablenotes}
* p-value $< 0.0001$ (compared to \emph{Baseline} results)
\end{tablenotes}
\end{threeparttable}
}
\end{table}
\begin{figure}[t]
    \centering   
    \includegraphics[width=0.9\textwidth]{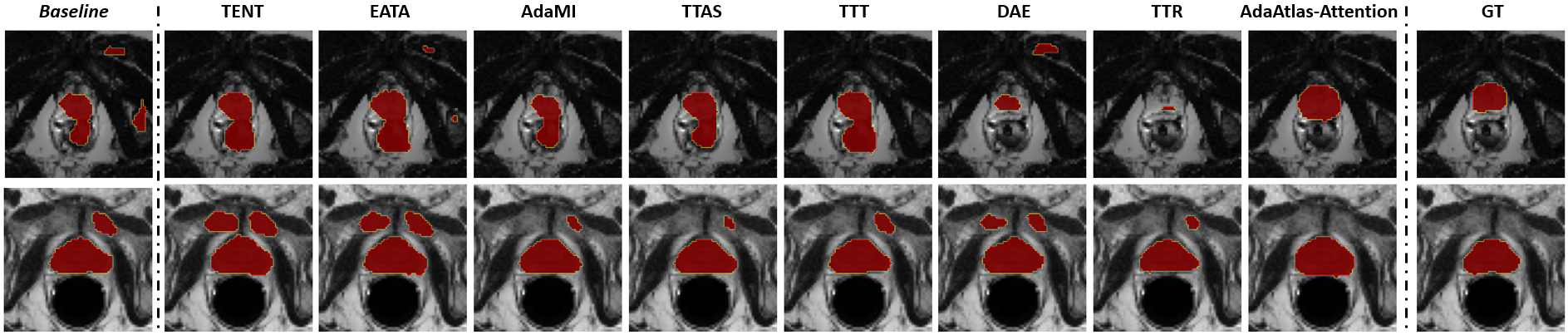}
    \caption{Visualization of segmentation results using \emph{Baseline} model (without TTA) and different TTA methods at test time using the same segmentation backbone (U-Net+Dual attention).  GT: ground truth.}
    \label{fig:comparison_visual}
\end{figure}
\noindent\textbf{Results.}
\label{sec:results}
Quantitative and qualitative results of prostate segmentation using U-Net or U-Net with dual attention blocks across six, unseen test domains are shown in Table~\ref{tab:prostate_TTA} and Fig.~\ref{fig:comparison_visual}. We can see that model adapted to target domains using different TTA methods consistently outperform the \emph{Baseline} model without TTA. Among all competitive methods, ours perform the best ($\sim $20 \% improvements) in both scenarios, indicating the superiority of using the high-level atlas-based shape prior to guiding TTA, which effectively cleans noisy predictions that are counter-intuitive in anatomical structures, see Fig~\ref{fig:comparison_visual}. Interestingly, we found TTR can completely fail in some cases, see the top row in Fig~\ref{fig:comparison_visual}. This can be because TTR highly depends on the quality of predicted segmentation from $\segNet_\theta$ for the refinement of the registration network, which can mislead the optimization process when the reference prediction is poor. Differently, we choose to use the reliable, domain-invariant atlas as supervision to refine the segmentation network instead, which is proved to be more robust. When comparing our AdaAtlas variants, AdaAtlas adapting the dual attention blocks (AdaAtlas-Attention) achieves higher improvements at test time compared to adapting the normalization blocks (AdaAtlas-Norm), especially on the challenging domain C. This verifies the benefits of increased flexibility using dual attention (channel+spatial) at test time. Although incorporating dual attention into U-Net can already bring a small target domain performance increase before adaptation (mean Dice score: 0.6610$\rightarrow$0.6770), refining them using our TTA method can bring much larger performance gains (0.6770$\rightarrow$0.8205). 
Besides prostate segmentation, we also conducted experiments on the atrial segmentation task on public multi-site MRI datasets and obtained similar findings, where our method works the best. Please see the supplementary material for more details.  

\noindent\textbf{Ablation study.}
We further performed two ablation studies to systematically analyze the contribution of two key components in our best-performing method (AdaAtlas-Attention), by replacing $\Loss_{atlas}$ with other TTA losses (incl. entropy loss $\Loss_{ent}$ used in \cite{wang2020tent}; the shape-moment-based loss $\Loss_{shape}$ used in the most competitive method TTAS \cite{bateson2022test}; the shape correction loss $\Loss_{DAE}$ used in DAE~\cite{Larrazabal_2019_MICCAI_DAE} or by restricting the adaptation blocks to be normalization only, channel attention only and spatial attention only, see Table ~\ref{tab:ablation}. Among all the variants, ours works the best. The success comes from a more detailed, reliable shape knowledge encoded in the atlas compared to vague, abstract shape statistics (moments) in $\Loss_{shape}$~\cite{bateson2022test}. While DAE-based shape correction~\cite{Larrazabal_2019_MICCAI_DAE} provides an estimated surrogate segmentation label for guidance, unlike ours, the supervision signal is conditioned on the predicted segmentation and can be very unstable and unreliable when the initial prediction is very poor and uninformative, e.g., totally missed segmentation during optimization.  
\begin{table}[t]
\centering
\caption{The effect of different TTA losses as well as the impact of adapting parts of the adaptation blocks.  All experiments are conducted with the same pretrained segmentation model (U-Net+Dual attention).}
\label{tab:ablation}
\resizebox{\textwidth}{!}
{
\setlength{\tabcolsep}{5pt}
\begin{tabular}{c|ccc|c||ccc|c}
\hline
                      & \multicolumn{4}{c||}{Different losses} & \multicolumn{4}{c}{Different adaptation blocks}                     \\ \hline
                      & $\Loss_{ent}$  & $\Loss_{shape}$&
                      $\Loss_{DAE}$  & \textbf{$\Loss_{atlas}$}\textbf{(ours)} & Norm & Channel only & Spatial only & \textbf{Dual attention(ours)} \\ \hline
Mean Dice score (A-F) & 0.6835   & 0.7893     &  0.7536 & 0.8205     & 0.8054 & 0.8103       & 0.8082       & 0.8205                               \\ \hline
\end{tabular}
}
\end{table}



\section{Discussion and conclusion}
We propose a novel, effective atlas-based TTA method (AdaAtlas) to distill high-level shape knowledge to better guide the adaptation process on out-of-domain scenarios, which is particularly useful for anatomical structure segmentation. We also investigate and compare different adaptation blocks (i.e., normalization and more flexible dual attention blocks) for an effective adaptation and find that adding dual attention blocks for adaptation (AdaAtlas-Attention) works the best. An interesting finding is that when using a less reliable entropy-based loss $\Loss_{ent}$ to adapt attention blocks, the model performance significantly drops ($0.8205\rightarrow 0.6835$, Table~\ref{tab:ablation}), which is even lower than using $\Loss_{ent}$ to adapt normalization blocks (0.6933 (TENT), Table~\ref{tab:prostate_TTA}). Our study uncovers the importance of building \emph{flexibility} in the design of a network for TTA and the \emph{reliability} of the guiding signal for robust TTA. One limitation of AdaAtlas is that it needs to train an additional atlas registration network at training time. This type of training modification requirement can be also found in other powerful TTA methods~\cite{sun2020test,sinclair2022atlas,karani2021test}. Our training dependency can be potentially removed by utilizing public atlases and existing powerful registration toolkits to make AdaAtlas a fully post-hoc method to adapt any existing segmentation network. It is also interesting to explore bi-level optimization for both segmentation and registration network, as well as dynamic learning rate~\cite{yang2022dltta} and extend it to a multi-atlas solution for further improvements.

\subsubsection{Acknowledgment:} This work was supported by ERC Advanced Grant \linebreak Deep4MI (884622) and the JADS programme at the UKRI Centre for Doctoral Training in Artificial Intelligence for Healthcare (EP/S023283/1).

\clearpage
\printbibliography
\appendix
\clearpage
\end{document}


\appendix

\beginsupplement
\begin{figure}[th]
    \centering   
    \includegraphics[width=0.5\textwidth]{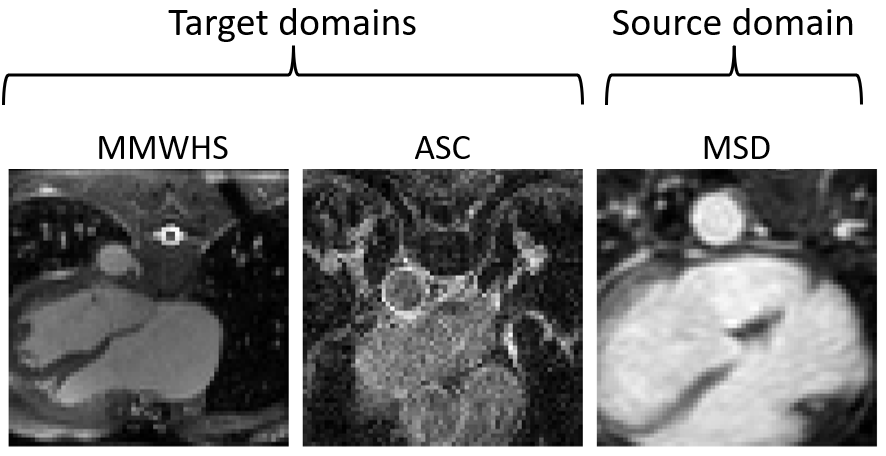}
    \caption{Visualization of atrial MRI datasets from different sites. Datasets MMWHS~\cite{zhuang2016multi} and ASC~\cite{xiong2021global} are target domains for testing. Dataset MSD~\cite{antonelli2022medical} is the training source domain.}
    \label{fig:heart_dataset_visual}
\end{figure}

\begin{table}[h]
\centering
\caption{Segmentation results of different test-time adaptation methods on atrial MRI unseen target domains, MMWHS~\cite{zhuang2016multi} and ASC~\cite{xiong2021global} with 20 and 54 subjects for testing respectively. MSD~\cite{antonelli2022medical} dataset is the training source domain for pre-training with 14 subjects. U-Net+Dual attention is the backbone for all the compared methods. Reported values are average Dice scores.}
\label{tab:cardiac_TTA}
\resizebox{1\textwidth}{!}
{%
\setlength{\tabcolsep}{5pt}
\begin{tabular}{c|c|c}
\hline
Method                   & MMWHS        & LA2018 \\ \hline
Baseline                 & 0.7657~\textcolor{gray}{(-)}       & 0.4989~\textcolor{gray}{(-)} \\ \hline
TENT                     & 0.7720~\textcolor{gray}{(+0.8\%)}       & 0.5276~\textcolor{gray}{(+5.8\%)} \\
EATA                     & 0.7784~\textcolor{gray}{(+1.7\%)}       & 0.5627~\textcolor{gray}{(+12.8\%)} \\
AdaMI                    & 0.7836~\textcolor{gray}{(+2.3\%)}       & 0.6091~\textcolor{gray}{(+22.1\%)} \\
TTAS                     & 0.7890~\textcolor{gray}{(+3.0\%)}       & 0.6316~\textcolor{gray}{(+26.6\%)} \\ \hline
TTT                      & 0.7675~\textcolor{gray}{(+0.2\%)}       & 0.5048~\textcolor{gray}{(+1.2\%)} \\
DAE                      & 0.7850~\textcolor{gray}{(+2.5\%)}       & 0.5977~\textcolor{gray}{(+19.8\%)} \\
TTR                      & 0.7792~\textcolor{gray}{(+1.8\%)}       & 0.5306~\textcolor{gray}{(+6.4\%)} \\ \hline
AdaAtlas-Norm(ours)      & 0.7961~\textcolor{gray}{(+4.0\%)}       & 0.7264~\textcolor{gray}{(+45.6\%)} \\
AdaAtlas-Attention(ours) & 0.8013~\textcolor{gray}{(+4.7\%)}       & 0.7488~\textcolor{gray}{(+50.1\%)} \\ \hline
\end{tabular}
}
\end{table}
\begin{figure}[t]
    \centering   
    \includegraphics[width=\textwidth]{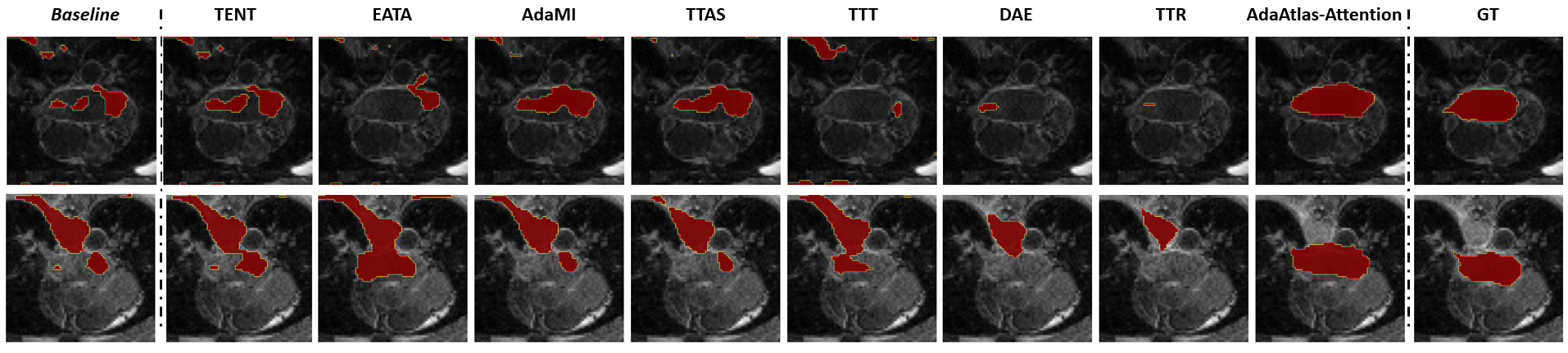}
    \caption{Visualization of atrial segmentation results using \emph{Baseline} model (w/o TTA) and different TTA methods at test time using the same segmentation backbone U-Net with dual attention blocks).  GT: ground truth.}
    \label{fig:comparison_visual_supple}
\end{figure}
\begin{figure}[t]
    \centering   
    \includegraphics[width=0.95\textwidth]{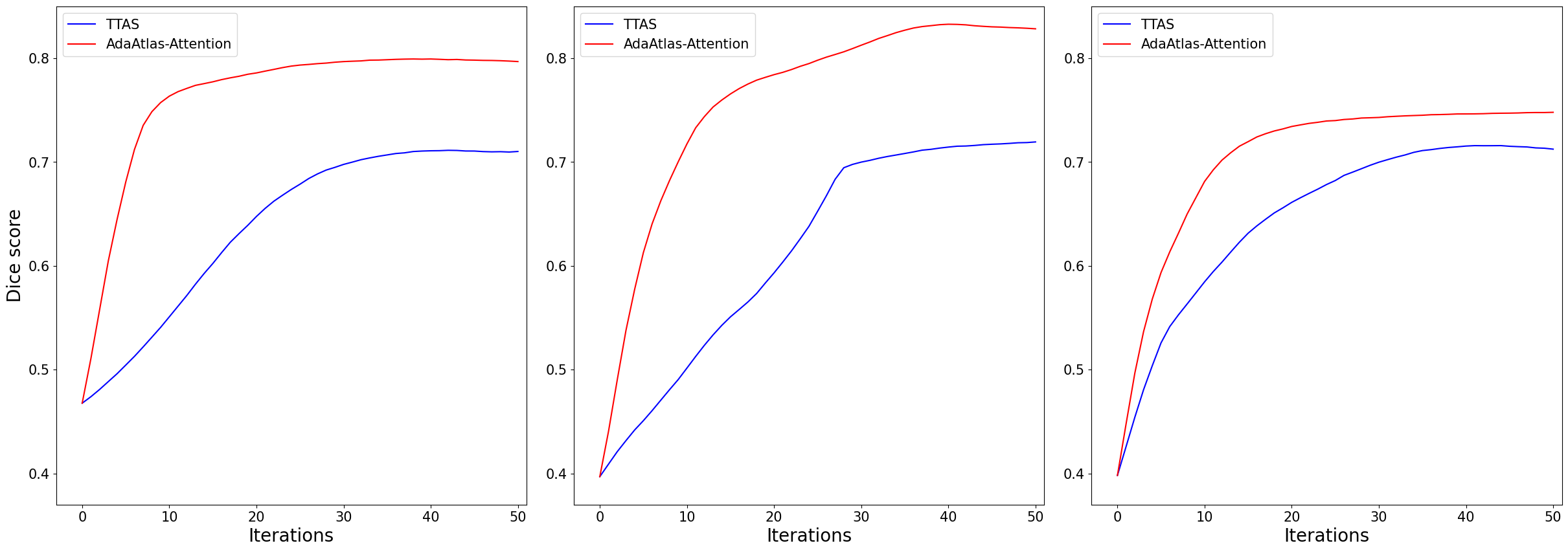}
    \caption{Dice curves of three different subjects from unseen target domains during test-time adaptation for our method \textbf{AdaAtlas-Attention} and the competitive method TTAS.}
    \label{fig:dice_curves}
\end{figure}

\printbibliography